\documentclass[final,5p,times,twocolumn]{elsarticle}

\usepackage{amssymb}
\usepackage{amsmath}

\usepackage[noend]{algorithmic}
\usepackage{algorithm}
\usepackage{amsmath,amsfonts}
\usepackage[inline]{enumitem}
\usepackage{mathtools}
\usepackage{multirow}
\usepackage{refcount}
\usepackage[caption=false]{subfig}
\usepackage{xcolor}

\newcommand{\union}{\textsc{Union}}
\newcommand{\find}{\textsc{Find}}
\newcommand{\RR}{$\mathit{RR}$}

\newcommand{\CALL}[2]{\textsc{#1}(#2)}

\journal{Journal of Parallel and Distributed Computing}

\begin{document}

\begin{frontmatter}



\title{Parallel Watershed Partitioning:\\
GPU-Based Hierarchical Image Segmentation}

\author[aua,oxford]{Varduhi Yeghiazaryan}
\author[aua,denovo]{Yeva Gabrielyan}
\author[oxford]{Irina Voiculescu}
\affiliation[aua]{organization={American University of Armenia},
            addressline={40 Marshal Baghramyan Ave},
            city={Yerevan},
            postcode={0019},
            country={Armenia}}
            
\affiliation[oxford]{organization={University of Oxford},
            addressline={Department of Computer Science, Wolfson Building, Parks Road},
            city={Oxford},
            postcode={OX1 3QD},
            country={UK}}
            
\affiliation[denovo]{organization={Denovo Sciences LLC},
            addressline={138/2 Verin Antarain Str},
            city={Yerevan},
            postcode={0009},
            country={Armenia}}



\begin{abstract}
Many image processing applications rely on partitioning an image into disjoint regions whose pixels are `similar.' 
The watershed and waterfall transforms are established mathematical morphology pixel clustering techniques.
They are both relevant to modern applications where groups of pixels are to be decided upon in one go, or where adjacency information is relevant.
We introduce three new parallel partitioning algorithms for GPUs. By repeatedly applying watershed algorithms, we produce waterfall results which form a hierarchy of partition regions
over an input image. Our watershed algorithms attain competitive execution times in both 2D and 3D, processing an 800 megavoxel image in less than 1.4~sec. 
We also show how to use this fully deterministic image partitioning as a pre-processing step to machine learning based semantic segmentation. This replaces the role of superpixel algorithms, and results in comparable accuracy and faster training times.
\end{abstract}




\end{frontmatter}



\section{Introduction}

Image partitioning was originally known in Computer Vision under the name of `image segmentation'. Nowadays, `image segmentation' often refers to semantic segmentation, where an object is segmented out of an image. The {\em fully deterministic} techniques proposed in this work refer to {\em partitioning an image} into disjoint regions of interest---with the ultimate downstream goal of improving segmentation or classification algorithms.

With a constant increase in the resolution of input into imaging algorithms, many techniques resort to processing multiple pixels in one go. This is achieved through downsampling the input image gradually through the depths of the layers of a machine learning network, or through grouping pixels together into superpixels~\cite{li2015superpixel,li2021superpixel}.
Pixel grouping is carried out as a one-off pre-processing step and merits particular attention due to the quality of its results influencing the downstream pipeline.



\begin{figure}[!htb]

    \centering
    \includegraphics[width=.45\linewidth]{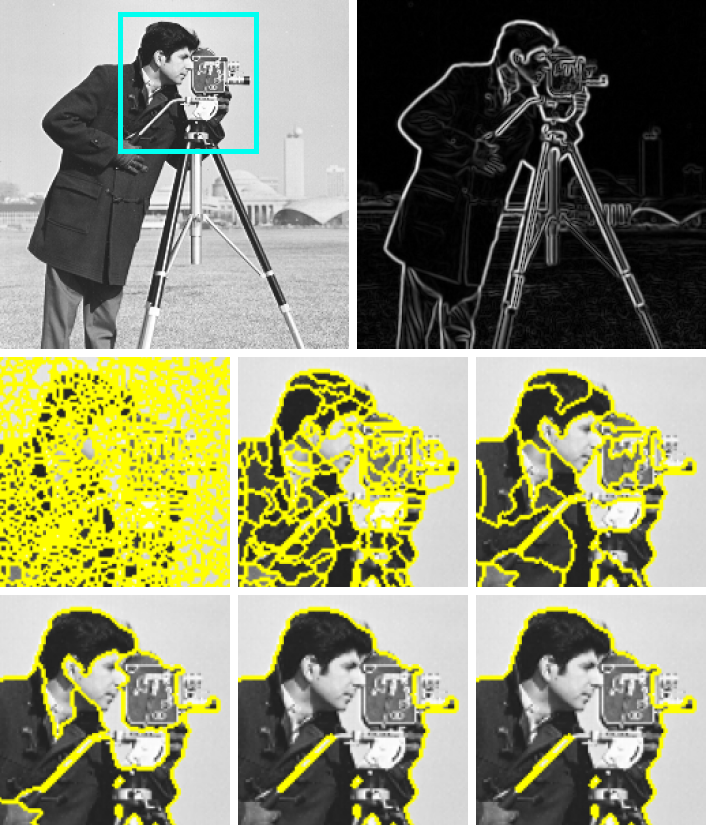}
    \includegraphics[width=.45\linewidth]{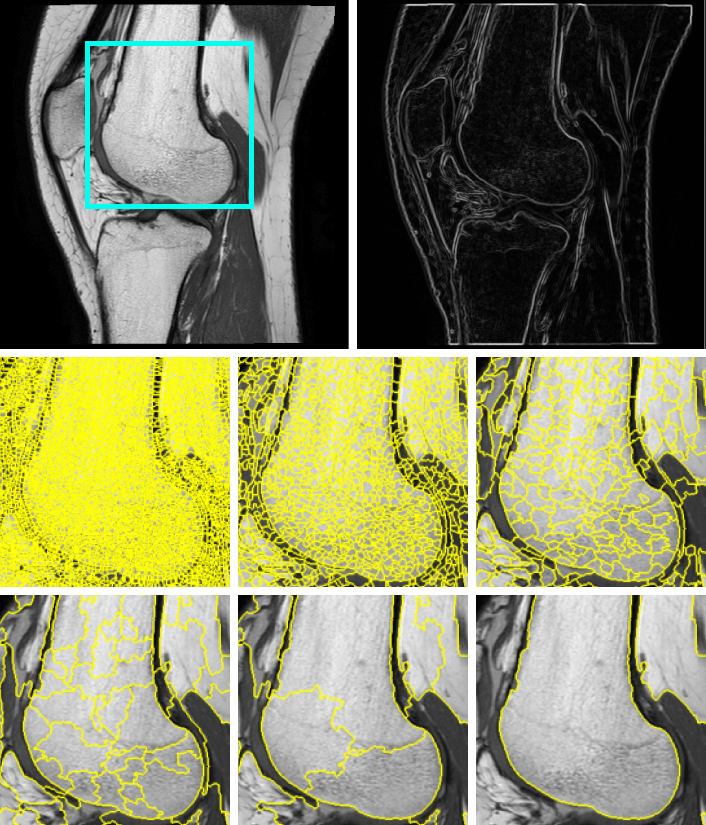}
    
\caption{Hierarchical output 
on two input images: cameraman and knee MRI slice. Top: raw greyscale inputs, gradient magnitude images (after Gaussian blurring); middle and bottom: cropped originals overlaid with watershed/waterfall lines in yellow for $\mathit{NL}{=}6$ hierarchical layers.}
\label{fig:qual}
\end{figure}

In the 1990s, image partitioning was synonymous with the watershed transform, which has been utilized for segmentation and feature extraction~\cite{tarhini2020detection,monteiro2019detecting,xu2020segmentation,tareef2018multi}. The watershed has also been used as a pre- or post-processing step in various deep learning~(DL) applications~\cite{das2019deep,elbatel2022seamless,eschweiler2019cnn,lux2019dic,jiang2019accurate}. 
However, watersheds have two main drawbacks: over-segmentation (regions are too small) and slow processing times (if executed sequentially). 

Our previous work~\cite{yeghiazaryan2018path,gabrielyan2022} aired three fast GPU watershed algorithms: PRW, APRUF and PRUF, which bring the watershed up to modern requirements. This paper builds on that previous work by speeding up and extending it to employ
watershed algorithms multiple times to achieve a hierarchical waterfall transformation, mitigating over-segmentation.

Our goal is to construct hierarchical segmentations (or partitions) of decreasing granularity, from input images of ultra-high resolution. This is achieved through deploying fast, parallel GPU-based watershed and waterfall transforms. These cluster pixels into equivalence classes through Union--Find, and then cluster these classes into larger partitions.
%
The steps are:
\begin{enumerate*}
    \item smooth the input image (minimally),
    \item construct a gradient magnitude image,
    \item apply the watershed transform, and
    \item repeatedly apply the waterfall transform.
\end{enumerate*} Fig.~\ref{fig:qual} illustrates output from this pipeline on two sample images: cameraman
and knee magnetic resonance imaging (MRI) slice. The segmented region boundaries, called watershed lines, in yellow are overlaid on the original greyscale images. 

The results of our experiments carried out on multiple datasets and GPUs demonstrate that the proposed algorithms outperform previous watershed algorithms for GPU execution in both effectiveness and efficiency. 
%
Furthermore, we show that the proposed algorithm for the hierarchical waterfall transform can be used as a replacement for superpixel algorithms in the pre-processing step of 
DL approaches to hyperspectral image classification applications (up to $3.9\times$ speedup).
This work highlights the versatility, relevance, and continued utility of parallel watershed/waterfall algorithms in Computer Vision.

The main contributions of this paper are as follows:
\begin{enumerate}[label=(\roman*)]
  \item three GPU watershed algorithms (PRW, PRUF, APRUF); earlier versions in our conference papers~\cite{yeghiazaryan2018path,gabrielyan2022};
  \item novel GPU waterfall that mitigates over-segmentation; 
  \item analysis of the effect of parameters (connectivity);
  \item Computer Vision applications of waterfall algorithm.
\end{enumerate}

\section{Related Work}
\label{sec:related}

The watershed is a greyscale image transformation from mathematical morphology that partitions the image into multiple small regions.
Images are interpreted as topographic terrain maps: white pixels correspond to the highest and black to the lowest altitudes. The watershed transform segments the image into disjoint regions called catchment areas, or basins. These regions are formed around regional minima in the topographic terrain. The boundary pixels separating the different catchment basins are called watershed lines, or watersheds. When the watershed transform is applied to the gradient magnitude image, the boundaries of catchment basins are placed at pixels with high gradient magnitude values (these correspond to boundaries of regions of interest).

A detailed discussion of watersheds is given in~\cite{roerdink2000}. 
%
%
Implementations of watersheds with distributed memory parallelism~\cite{swiercz2010fast,wu2012parallelization}, shared memory parallelism~\cite{wagner2009parallel,van2011towards} and  GPUs~\cite{kauffmann2008,korbes2011,hucko2012streamed,quesada2012} have also been explored. Watersheds on GPUs have been the subject of recent research due to the increasing demand for faster image processing. Initially GPU implementations of the watershed used `flooding'~\cite{pan2008implementation,kauffmann2008,kauffmann2010seeded,wagner2010parallel}, but hybrid~\cite{vitor2009,korbes2009analysis} (named DW in the experimental comparison below) and parallel depth-first watershed 
algorithms~\cite{vitor2012analysis,korbes2011,quesada2012,quesada2013} have also been proposed.

Computer Vision applications of the watershed~\cite{derivaux2010supervised,tarhini2020detection,monteiro2019detecting,xu2020segmentation,tareef2018multi} include image segmentation, object boundary detection, image restoration. It is also commonly used as a pre- or post-processing step for various applications working with microscopic images~\cite{eschweiler2019cnn,lux2019dic}, micro-CT images~\cite{xu2020segmentation}. 

The waterfall~\cite{beucher1990segmentation,beucher1994} is a hierarchical segmentation transform based on the watershed. It is a merging technique which addresses over-segmentation. The main idea of waterfall is that catchment basins from the watershed get turned into new plateaux and the watershed is re-applied. 
The resulting image contains a reduced number of regional minima. 
This recursive process
produces a series of image partitionings with a monotonically decreasing number of catchment basins, until a single
region corresponds to the whole image. Aside from the waterfall that can be directly applied on an image, a number of graph-based waterfall algorithms have been proposed for fast sequential implementation~\cite{marcotegui2005,golodetz2014two}.

In order to mitigate the shortcomings of classical waterfall (early removal of important contours from the hierarchy and the difficulty of finding a `good' depth of hierarchy) three modifications of the waterfall transform have been proposed. The standard and the P algorithms~\cite{beucher2009p} replace the hidden parameter called gain with an explicit one. Enhanced waterfall transform~\cite{beucher2013towards} re-introduces selected contours into the watershed results of each level of the hierarchical segmentation based on specific rules. An intuitive application of the classical waterfall transform is the construction of image partition forests (IPFs)~\cite{golodetz2010zipping,golodetz2017simpler}.




\section{Proposed Watershed Algorithms}
\label{sec:watershed}

\begin{figure}[!htb]
	\centering
	\includegraphics[width=.9\linewidth]{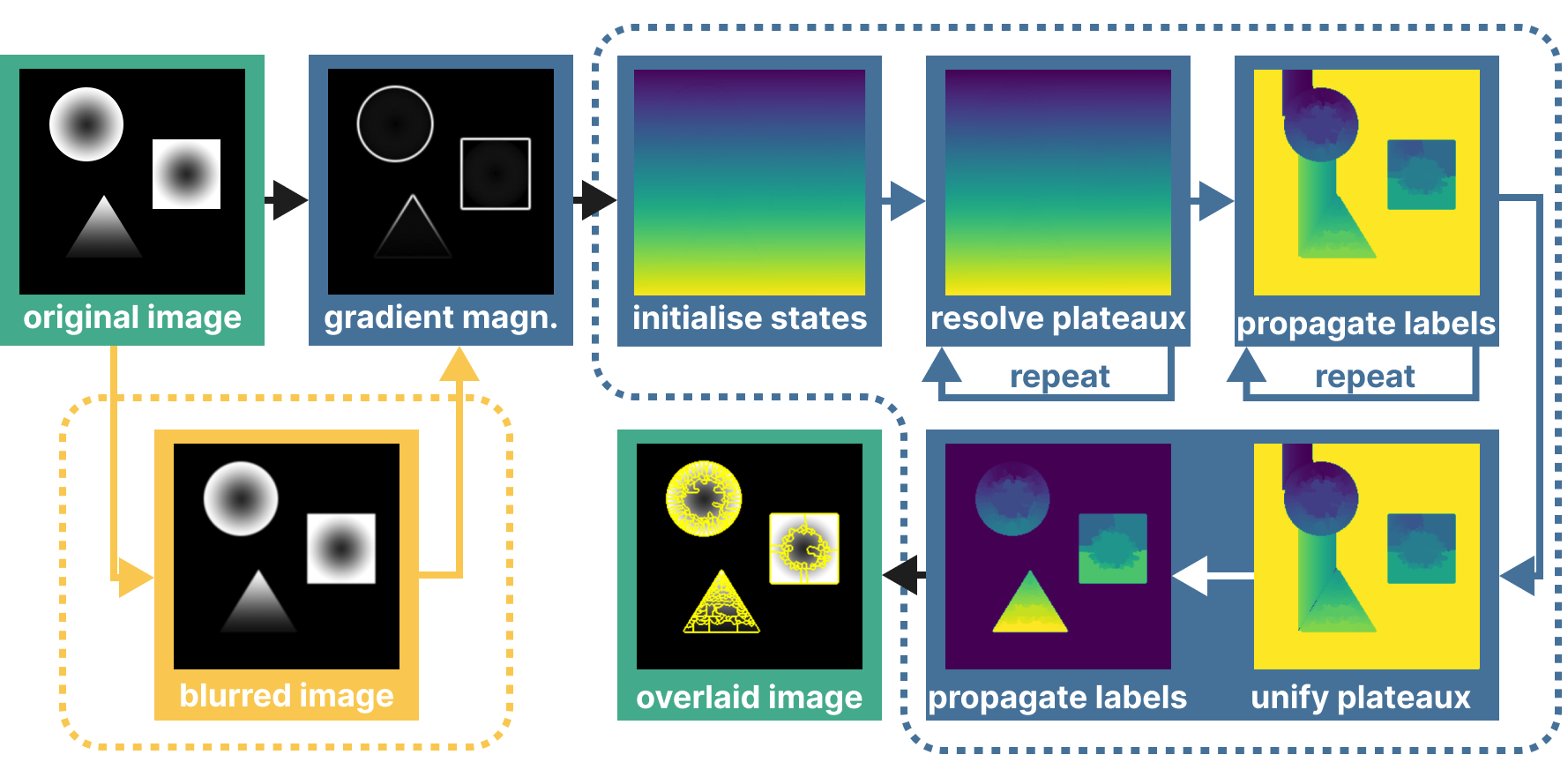}
	\caption{PRUF pipeline inside blue dashed outline. Each of steps I--III corresponds to one CUDA kernel: I executed once, II and III repeated iteratively. Step IV is achieved through two CUDA kernels, each executed once. Input image smoothing and use of gradient magnitude image are optional.}
\label{fig:watershed_flow}
\end{figure}

We present three parallel algorithms for the watershed transform: path reducing watershed (PRW), path reducing \union--\find\ watershed (PRUF) and alternative PRUF watershed (APRUF). In the next section we show that, with the addition of two CUDA kernels, it is possible to produce a waterfall algorithm from each of these watershed algorithms.

Each one of our parallel watersheds consists of four major steps:
I: initialise pixel states and labels,
II: resolve non-minimal plateaux,
III: propagate labels, or reduce paths, and
IV: merge minimal plateau labels.
This process belongs to the class of topographic-distance-based watershed transforms; it exploits paths of steepest descent. The four steps can be alternatively viewed as: I:~initialise paths of steepest descent; II:~resolve paths on non-minimal plateaux; III:~reduce paths into pointers to local minima; IV:~add extra paths inside minimal plateaux to achieve a single pointer target, i.e.\ label, for each catchment basin.

Figure~\ref{fig:watershed_flow} illustrates the general workflow for PRUF. Details of PRW and APRUF are described later in the text.

\subsection{Path-Reducing, \union--\find\ Watershed}

%
%
\begin{algorithm}[!htb]
\footnotesize
\caption{Path reducing, \union--\find\ watershed}
\label{alg:pruf}
\begin{algorithmic}[1]
\FORALL[Step I: initialise labels \& states] {$p \in P$ in parallel}
	\STATE $q \coloneq \max\{r \in N(p) \;|\; \forall n \in N(p), I(r) \le I(n)\}$
	\IF{$I(q) < I(p)$}
		\STATE $L(p) \coloneq q;~ S(p) \coloneq 0$
	\ELSIF{$I(q) > I(p)$}
		\STATE $L(p) \coloneq p;~ S(p) \coloneq 1$
	\ELSIF{$q > p$}
		\STATE $L(p) \coloneq q;~ S(p) \coloneq 2$
	\ELSIF{$q < p$}
		\STATE $L(p) \coloneq p;~ S(p) \coloneq 3$
	\ENDIF
\ENDFOR
\REPEAT[Step II: resolve non-minimal plateaux]
	\FORALL{$p \in P$ in parallel}
		\IF{$S(p) \ge 2$ \textbf{and} $\exists q \in N(p), \; S(q)=0 \land I(q)=I(p)$}
			\STATE $L(p) \coloneq q;~ S'(p) \coloneq 0$
		\ELSE
			\STATE $S'(p) \coloneq S(p)$
		\ENDIF
	\ENDFOR
	\STATE $swap(S, S')$
\UNTIL{$\neg change$}
\REPEAT[Step III: propagate labels, or reduce paths]
	\FORALL{$p \in P$ in parallel}
		\FOR{$i \coloneq 1,RR$ \textbf{and} $L(p) \ne L(L(p))$}
			\STATE $L(p) \coloneq L(L(p))$
		\ENDFOR
	\ENDFOR
\UNTIL{$\neg change$} \label{alg:stepS3}
\FORALL[Step IV: merge min plateau labels] {$p \in P$ in parallel}
	\IF{$S(p) \ge 2$}
		\FORALL{$q \in N(p), q > p, S(q) \ge 2$}
			\STATE \CALL{Union}{$L, p, q$}
		\ENDFOR
	\ENDIF
\ENDFOR
\FORALL{$p \in P$ in parallel}
	\STATE $L(p) \coloneq $ \CALL{Find}{$L, p$}
\ENDFOR
\end{algorithmic}
\end{algorithm}

Algorithm~\ref{alg:pruf} details PRUF.
Let $P$ be the set of all pixel positions in the image, and for every $p \in P$ let the local neighbourhood $N(p)$ be the set of pixels in the image that are a unit distance away from~$p$. The obvious choices are either Moore neighbourhood with 8 pixels in 2D and 26 voxels in 3D, or von Neumann neighbourhood with 4 pixels in 2D and 6 voxels in 3D, both reported hereafter.

\begin{figure}[!htb]
	\centering

\subfloat[{\scriptsize $6{\times}2$ greyscale input with expected watershed line}]{
\includegraphics[width=.8\linewidth]{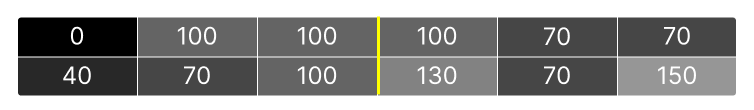}
}

\subfloat[{\scriptsize step I: initialise states (colours) and labels (arrows)}]{
\includegraphics[width=.8\linewidth]{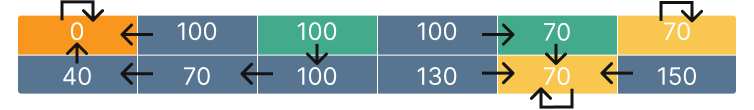}
}

\subfloat[{\scriptsize step II: update non-minimal plateau states and labels}]{
\includegraphics[width=.8\linewidth]{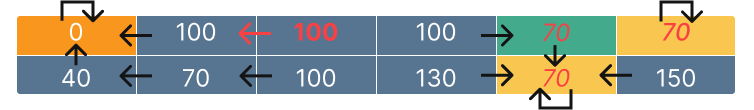}
}

\subfloat[{\scriptsize step III: propagate minima labels uphill to all pixels}]{
\includegraphics[width=.8\linewidth]{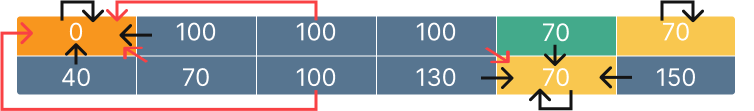}
}

\subfloat[{\scriptsize step IV: merge minimal plateau labels}]{
\includegraphics[width=.8\linewidth]{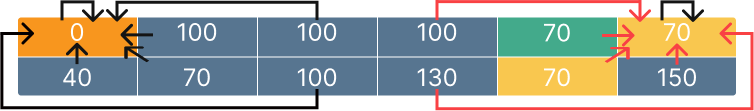}
}

	\caption{Simple $6{\times}2$ greyscale input image illustrates our path reducing watershed algorithms (4-pixel neighbourhood). State and label updates are highlighted in \textcolor[HTML]{F94144}{\bf red boldface}; \textcolor[HTML]{F94144}{\it red italics} indicate that an update was considered but not performed. Pixel states are colour-coded: \colorbox[HTML]{577590}{\phantom{A}} state 0, \colorbox[HTML]{F8961E}{\phantom{A}} state 1, \colorbox[HTML]{43AA8B}{\phantom{A}} state 2, and \colorbox[HTML]{F9C74F}{\phantom{A}} state 3.}
 
\label{fig:example}
\end{figure}

During step I: initialisation, all pixels are classified into four groups, called states, using the image values of the neighbouring pixels. Our algorithm requires that all pixels in the image be ordered; our implementation uses row-major order. Let $q \in N(p)$ be the pixel neighbouring $p$ that has the smallest input image value. If the minimum value occurs at multiple neighbours, let $q$ be the last in the fixed order among all pixels in that neighbourhood that have the minimum image value:
\begin{equation}
q = \max\{r \in N(p) \;|\; \forall n \in N(p), I(r) \le I(n)\},
\end{equation}
where $I(p)$ is the image value at pixel $p \in P$.
 Fig.~\ref{fig:example} details the steps of the watershed on a small example.

\begin{itemize}
    \item If $I(q) < I(p)$, then the direction of steepest descent from $p$ is $q$. The initial label $L$ of $p$ is set to $q$; later it will borrow its eventual label from $q$. Thus, pixel $p$ is assigned to be in state 0 to indicate it has a lower neighbour ($S(p)=0$).
    \item If $I(q) > I(p)$, then $p$ is a local minimum. We set $L(p) = p$, so that later this value can be used to label the whole catchment basin. The set of local minima is indicated by their states fixed at 1.
    \item Lastly, if $I(q) = I(p)$, then $p$ is inside a plateau. We differentiate such pixels into two states:
        \begin{itemize}
            \item a plateau pixel that has at least one equal neighbour (row-major order) after it ($S(p)=2$) and
            \item a plateau pixel that have equal neighbours (row-major order) only before it ($S(p)=3$).
        \end{itemize}
        If the plateau is eventually, after step II, categorised as minimal, state-3 plateau pixels will lend their label to state-2 plateau pixels. Hence, $L(p) = q$ if $S(p)=2$ and $L(p)=p$ forms a self-loop if $S(p)=3$.
\end{itemize}

Step II resolves non-minimal plateaux. It propagates labels inwards into a non-minimal plateau from its boundaries. For every state-2 or state-3 pixel $p$, $S(p) \ge 2$, if $\exists q \in N(p)$ such that $S(q)=0$ and if $I(q) = I(p)$, then $S(p)$ is reset to 0 and $L(p)$ is updated to equal $q$. We repeat this process iteratively until convergence with no state updates. The states are buffered during iterations and swapped between iterations to achieve consistent propagation of labels into non-minimal plateaux (see further discussion in Sec.~\ref{ssec:async}).

After step II the pixel states can be seen in a new way, slightly different from the initial interpretation. State~0 corresponds to pixels with a lower neighbour and non-minimal plateau pixels. States 2 and 3 only mark pixels within minimal plateaux. Since labels point to either a neighboring pixel or themselves and are acyclic, they can be followed recursively to create directed paths between pixels. The labels of all state-0 pixels form paths leading to minima, specifically state-1 or state-3 pixels. These steepest descent paths define different catchment basins. Additionally, within minimal plateaus, the labels of state-2 pixels create further paths to state-3 pixels. Therefore, any label path starting from any pixel will ultimately lead to a state-1 or state-3 pixel, where the label forms a self-loop.

\begin{figure}[!htb]
	\centering
	\includegraphics[width=.8\linewidth]{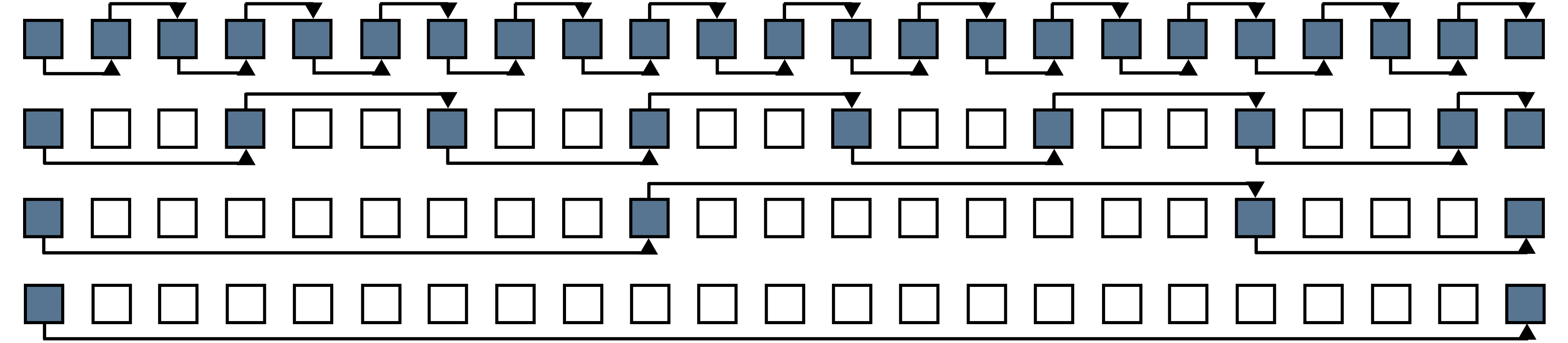}
\caption{Length 22 path with \RR${}=2$, in only $\lceil log_3 22 \rceil = 3$ iterations.}
\label{fig:red}
\end{figure}

In step III, all label paths are reduced into direct pointers from any image pixel into its corresponding catchment basin minimum state-1 or state-3 pixel. This process is performed independently and in parallel for each pixel, following the label pointers until a self-loop is encountered. However, this procedure requires time proportional to the length of the longest label path, making the worst-case time complexity linear relative to the image size. To reduce this complexity to logarithmic, we synchronize all labels after a fixed number of updates and repeat the process iteratively. This fixed number of updates between synchronizations is called \emph{reduction rate} ($RR$). The label of each pixel $p$ is updated to its label's label \RR\ times; the new label value is recorded instead of the original $L(p)$ and made available to other pixels to read. This iterative process continues until no further label changes occur. The overall number of operations is $O(RR {\times} \lceil \log_{RR+1} Len \rceil) = O(\lceil \log_{RR+1} Len \rceil) \le O(\log_{RR+1} |P|)$, where \RR\ is the reduction rate, $Len$ is the length of the longest path, $|P|$ is the image size. Figure~\ref{fig:red} demonstrates path reduction.

\begin{figure}[!htb]
	\centering
	\includegraphics[width=.6\linewidth]{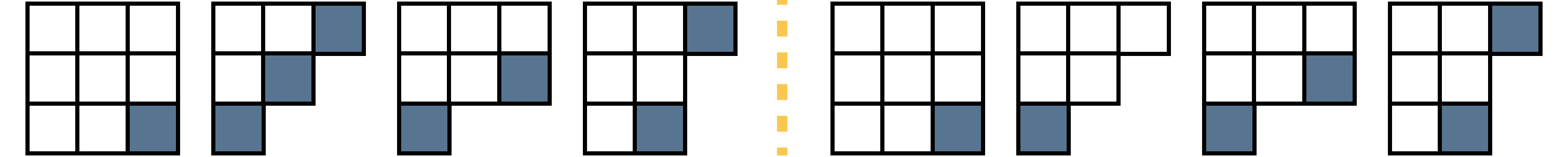}
	\caption{Minimal plateau shapes with state-3 pixels highlighted in a blue shade. 4-pixel neighbourhood on the left; 8-pixel on the right.}
\label{fig:minplat}
\end{figure}

For a 1D image, the label array produced after steps I-III of the algorithm constitutes a valid watershed result. This is because each minimal plateau in a 1D image forms an interval, and consequently, the only state-3 pixel within such a plateau is the rightmost pixel. As a consequence, after step III, each catchment basin in the label array has a single value. Still, this property does not hold for higher-dimensional images. Based on the shape of a minimal plateau, there may be several state-3 pixels in it. Figure~\ref{fig:minplat} shows a few minimal plateau shapes with state-3 pixels highlighted.

Step IV addresses the problem of multiple labels for minimal plateaux by merging those labels. Once steps I--III are complete, the pixel labels can be viewed as a \emph{disjoint-set} data structure. A disjoint set of pixels is formed for each one label, and all set elements point directly to the set representative. One or more disjoint sets correspond to every catchment basin in the watershed partitioning. When several labels, i.e.\ disjoint sets, overlap within the same catchment basin, it indicates that the basin has a plateau minimum containing multiple state-3 pixels (which can occur in 2D or higher-dimensional images). To address this, we use a parallel \union--\find\ implementation~\cite{allegretti2020} to merge all disjoint sets that represent the same catchment basin. Step IV performs a \union\ whenever it detects neighbouring minimal-plateau pixels with varying labels, thus merging the different labels within the same catchment basin. Finally, it updates each pixel's label to the index, or label, of the set representative by executing a \find operation.

When we introduced PRW~\cite{yeghiazaryan2018path} we showed 
it was significantly faster than existing approaches. It differs from PRUF only in step IV. The pseudocode for that step is presented in Alg.~\ref{alg:prw}. A further improvement over~\cite{yeghiazaryan2018path} and~\cite{gabrielyan2022} is to consider only half of the neighbourhood in step IV ($q > p$ on line 26 of Alg.~\ref{alg:pruf} and line 27 of Alg.~\ref{alg:prw}).

\begin{algorithm}[!htb]
\footnotesize
\caption{Path reducing watershed}
\label{alg:prw}
\begin{algorithmic}[1]
\setcounterref{ALC@line}{alg:stepS3}
\item[] \COMMENT{Steps I--III are the same as in Alg.~\ref{alg:pruf}}
\REPEAT[Step IV: merge minimal plateau labels]
	\FORALL{$p \in P$ in parallel}
		\IF{$S(p) \ge 2$}
			\FORALL{$q \in N(p), q > p, S(q) \ge 2$}
				\STATE $lp \coloneq L(p);~ lq \coloneq L(q)$
				\WHILE{$L(lp) \ne L(lq)$}
					\STATE $L(lp) \coloneq \min(L(lp), L(lq))$
					\STATE $L(lq) \coloneq \min(L(lp), L(lq))$
				\ENDWHILE
			\ENDFOR
		\ENDIF
		\WHILE{$L(p) \ne L(L(p))$}
			\STATE $L(p) \coloneq L(L(p))$
		\ENDWHILE
	\ENDFOR
\UNTIL{$\neg change$}
\end{algorithmic}
\end{algorithm}

For adjacent pixels $p, q$ within a minimal plateau, $lp = L(p), lq = L(q)$ are the representative labels of the corresponding sets. If $lp \ne lq$, then we merge the two sets by setting the smaller label as the parent pointer of the larger label. Since our procedure is designed for parallel implementation, more than two pixel sets can be merged simultaneously. This check and subsequent label update are performed repeatedly until $L(lp)$ equals $L(lq)$. A label path reduction loop is then carried out for all pixels to convert label paths into direct pointers to representative labels. This merging and reduction process continues iteratively until no further label changes occur. Upon completing step IV, each catchment basin will have a single label value, and for each pixel $p$, $L(p)$ indicates the catchment basin to which $p$ belongs.

\subsection{Alternative Path-Reducing, \union--\find\ Watershed}

A slightly revised version of the PRUF algorithm, known as APRUF, has been developed for relatively \emph{smaller} images (typically those under 10 megapixels/megavoxels). It alters the implementation of step III by substituting lines 19--23 in Alg.~\ref{alg:pruf} with lines 28--29. In smaller images, catchment basins are generally smaller, resulting in shorter label paths to minima. While PRUF requires global synchronization to make updated labels visible to other threads, APRUF improves execution time by eliminating this global synchronization: labels are independently acquired by each thread.


\subsection{Synchronous Implementation for the GPU}

Using the CUDA Toolkit by NVIDIA, we have produced three different implementations of our PRUF watershed in Alg.~\ref{alg:pruf}. The initial and the most basic implementation of PRUF is described first; referred to as \emph{synchronous} PRUF (PRUF\textsubscript{sync}). The remaining two implementations are discussed in Sec.~\ref{ssec:async}. Each of the three algorithm variants assigns a dedicated thread to each pixel in the image.

PRUF\textsubscript{sync} relies on a host code function and five CUDA kernels: one for each step of the process, plus an additional one for step IV. A kernel is a function that runs on numerous threads on the GPU in parallel. The five kernels in PRUF\textsubscript{sync} implement lines 1--10, 12--16, 20--22, 24--27, and 28--29 in Alg.~\ref{alg:pruf}, respectively.


The most expensive step of PRUF (for any implementation), and PRUF\textsubscript{sync} in particular, is the resolution of non-minimal plateaux. Unlike steps I and III--IV, the number of iterations of the code
in lines 11--18 in Alg.~\ref{alg:pruf} may be linear in the image size in the worst case. For instance, in the case of $(2n+2) {\times} 1$ example
\[
75 \quad \underbrace{89 \quad 89 \quad 89 \quad \dots \quad 89 \quad 89}_{2n} \quad 81,
\]
then it requires $n$ iterations to complete step II of PRUF\textsubscript{sync} due to the costly global inter-block synchronization that occurs after each state update. Pixel states are initialised in step I as
\[
1 \quad 0 \quad \underbrace{2 \quad 2 \quad \dots \quad 2}_{2n-2} \quad 0 \quad 1;
\]
after $n-2$ iterations of step II we have
\[
1 \quad \underbrace{0 \quad 0 \quad \dots \quad 0}_{n-1} \quad 2 \quad 2 \quad \underbrace{0 \quad 0 \quad \dots \quad 0}_{n-1} \quad 1.
\]
During the $(n-1)^\text{th}$ iteration, the remaining 2-states are converted to 0, and the final  $n^\text{th}$ iteration, which does not introduce any changes, completes the resolution of non-minimal plateaux.

Moreover, because state update for the plateau pixels in lines 13--16 of Alg.~\ref{alg:pruf} occur in parallel and the order of update completion is non-deterministic, we require the buffer state memory $S'$ to ensure correctness. The buffer also ensures a fair distribution of non-minimal plateaux among neighboring catchment basins. For the given example at $n=5$, the label values after step II will be
\[
1 \leftarrow 0 \leftarrow 0 \leftarrow 0 \leftarrow 0 \leftarrow 0 \phantom{{}\leftarrow{}} 0 \rightarrow 0 \rightarrow 0 \rightarrow 0 \rightarrow 0 \rightarrow 1,
\]
where the numbers represent pixel states and the arrows indicate the labels.

Therefore, the sizes of non-minimal plateaux within an image is anticipated to significantly impact the execution time of parallel watershed processing. Such trends are documented in existing literature~\cite{quesada2013,korbes2011}; our results in Sec.~\ref{sec:results} corroborate this assumption.

\subsection{Block-Asynchronous and Balanced Versions}
\label{ssec:async}

Alongside PRUF\textsubscript{sync} we introduce two additional PRUF watershed implementations designed to enhance execution speed. These implementations leverage CUDA’s intra-block synchronization mechanisms. Block-level shared memory offers quicker access to data shared among threads, and in-block synchronization is less costly than global synchronization. By utilizing these block-level tools, we aim to minimize the number of global iterations required for resolving non-minimal plateaux.

The \emph{block-asynchronous} PRUF watershed (PRUF\textsubscript{async}) is different from PRUF\textsubscript{sync} in the second step of the algorithm.
Its step II CUDA kernel uses block-level synchronisation to perform multiple state updates. 

The non-minimal plateau resolution kernel of PRUF\textsubscript{async} begins by loading data into shared and local memory. Because each pixel requires access to both the input image and the state values of neighboring pixels to update its own state, this information is stored in shared memory. Each pixel $p$ is tasked with copying its $I(p)$ and $S(p)$ values into the shared memory. Additionally, pixels located at the block's boundary also copy the values of their neighbors outside the block. 

Each thread block operates with its own shared memory space, and there is no communication between different blocks during kernel execution. Data updates from one block become available to threads in other blocks only after being written to global memory and upon completion of kernel execution, which signifies global synchronization points. Consequently, the shared memory values for the additional band of pixels remain unchanged during kernel execution (they are updated only within their respective block) and retain the values from the last global synchronization point. Information is transferred from one block to another once per global iteration.

Our third variant, \emph{balanced} block-asynchronous PRUF (PRUF\textsubscript{bal}), combines block-level synchronization, similar to PRUF\textsubscript{async}, with the fair watershed lines positioning of, as PRUF\textsubscript{sync}. We expand the 0--3 state range to include negative numbers: when $S(p) < 0$ then value $|S(p)|$ indicates the distance of a non-minimal plateau pixel from the plateau boundaries, reflecting the number of state update operations required for information to reach $p$. Alg.~\ref{alg:bal} outlines the non-minimal plateau resolution step (kernel) for PRUF\textsubscript{bal}.

\begin{algorithm}[!htb]
\footnotesize
\caption{Non-minimal plateau resolution, PRUF\textsubscript{bal}}
\label{alg:bal}
\begin{algorithmic}[1]
\STATE $GlobalToLocal(L)$ \COMMENT{Copy label from global to local memory}
\STATE $GlobalToShared(I)$ \COMMENT{Copy image from global to shared}
\STATE $GlobalToShared(S)$ \COMMENT{Copy state from global to shared}
\REPEAT
	\STATE $SyncThreads$ \COMMENT{Synchronise threads in block}
	\STATE $SharedToLocal(S)$ \COMMENT{Copy state from shared to local}
	\IF{($S(p) \ge 2$ \textbf{and} $\exists q \in N(p),\; S(q) \le 0 \land I(q)=I(p)$)
	\;\textbf{or}\\\; ($\exists q \in N(p),\; S(p) + 1 < S(q) \le 0 \land I(q) = I(p)$)}
		\STATE $L(p) \coloneq q;~ S(p) \coloneq S(q)-1$ 
	\ENDIF
	\STATE $SyncThreads$ \COMMENT{Synchronise threads in block}
	\STATE $LocalToShared(S)$ \COMMENT{Copy updated state to shared memory}
\UNTIL{$\neg blockchange$}
\STATE $LocalToGlobal(L)$ \COMMENT{Copy updated label to global memory}
\STATE $SharedToGlobal(S')$ \COMMENT{Copy updated shared state to global}
\end{algorithmic}
\end{algorithm}

After a non-minimal-plateau pixel's state changes from 2 or 3 to a negative number, it can be updated iteratively to larger (but still negative) values during subsequent in-block iterations or following global synchronization points. The resolution step is deemed complete only when no additional improvements can be made to the states of the non-minimal plateau pixels. In the case of the simple $14 {\times} 1$ pixel example

\noindent
\begin{tabular}{*{14}{p{6pt}}}
75 &
89 & 89 & 89 & 89 & 89 & 89 &
89 & 89 & 89 & 89 & 89 & 89 &
81
\end{tabular}

\noindent
assuming block size 3, the first global iteration of step II produces states
\[
\underbracket[1pt]{1 \;\; \phantom{+}0 \;\; {-1}} 
\;\;
\phantom{+}\underbracket[1pt]{2 \;\; \phantom{+}2 \;\; \phantom{+}2} 
\;\; 
\phantom{+}\underbracket[1pt]{2 \;\; \phantom{+}2 \;\; \phantom{+}2} 
\;\; 
\underbracket[1pt]{{-3} \;\; {-2} \;\; {-1}} 
\;\; 
\phantom{+}\underbracket[1pt]{0 \;\; \phantom{+}1}.
\]
Following the second global iteration, we get
\[
\underbracket[1pt]{1 \;\; \phantom{+}0 \;\; {-1}} \;\; \underbracket[1pt]{{-2} \;\; {-3} \;\; {-4}} \;\; \underbracket[1pt]{{-6} \;\; {-5} \;\; {-4}} \;\; \underbracket[1pt]{{-3} \;\; {-2} \;\; {-1}} \;\; \phantom{+}\underbracket[1pt]{0 \;\; \phantom{+}1};
\]

and this is corrected after iteration 3
\[
\underbracket[1pt]{1 \;\; \phantom{+}0 \;\; {-1}} \;\; \underbracket[1pt]{{-2} \;\; {-3} \;\; {-4}} \;\; \underbracket[1pt]{\mathbf{-5} \;\; {-5} \;\; {-4}} \;\; \underbracket[1pt]{{-3} \;\; {-2} \;\; {-1}} \;\; \phantom{+}\underbracket[1pt]{0 \;\; \phantom{+}1}.
\]



PRUF\textsubscript{bal} is slightly slower than PRUF\textsubscript{async} due to (a) additional state updates and (b) extra global iterations of resolution. Overall, PRUF\textsubscript{bal} shows a large runtime improvement upon PRUF\textsubscript{sync}, especially on images with large plateaux.

Similar implementation variants can be considered for PRW and APRUF, producing PRW\textsubscript{sync}, PRW\textsubscript{async}, PRW\textsubscript{bal}, APRUF\textsubscript{sync}, APRUF\textsubscript{async}, and APRUF\textsubscript{bal}.
The main parameters of the path reducing, \union--\find\ watershed affecting execution times are the thread block size and \RR.

\begin{figure}
	\centering
	\includegraphics[width=0.7\linewidth]{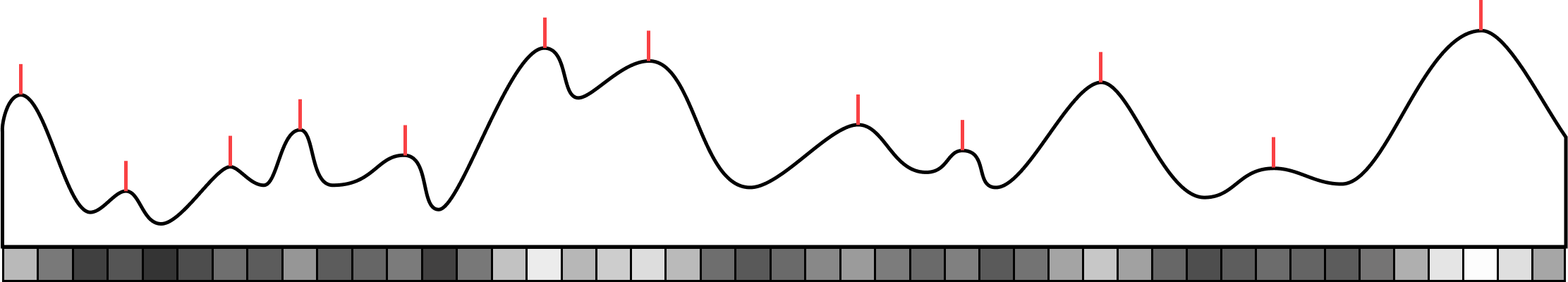}\\
    \vspace{5pt}
	\includegraphics[width=0.7\linewidth]{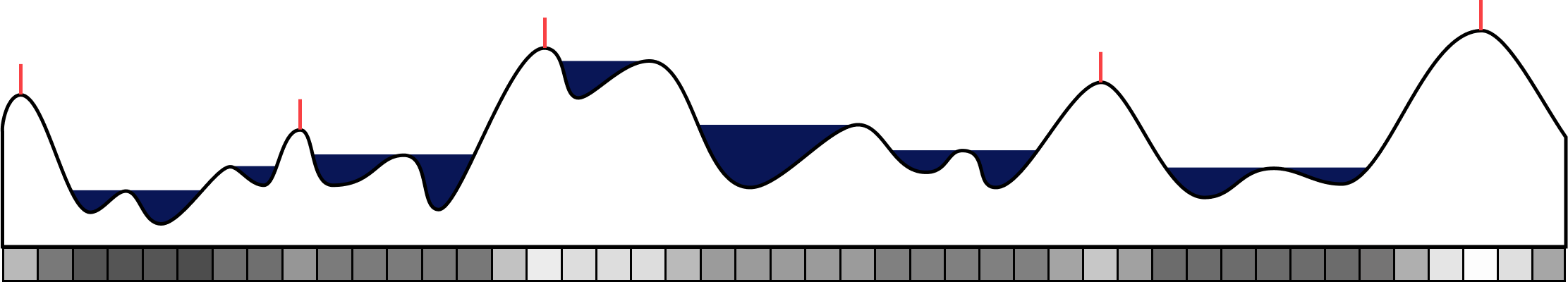}\\
    \vspace{5pt}
	\includegraphics[width=0.7\linewidth]{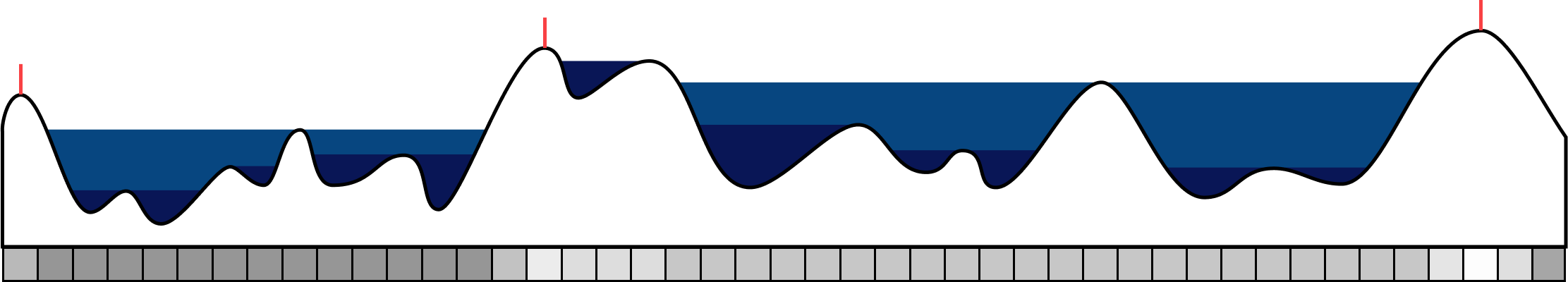}\\
    \vspace{5pt}
	\includegraphics[width=0.7\linewidth]{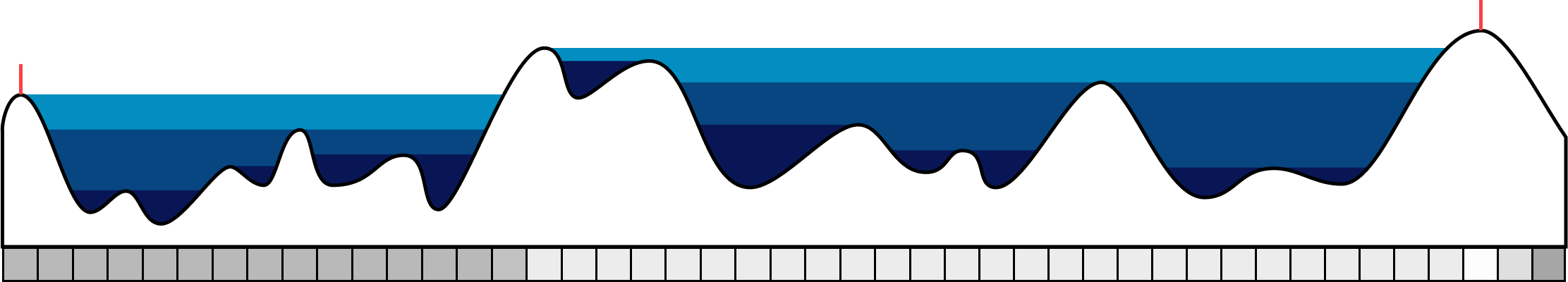}
 \caption{Watershed transform of 1D image and three consecutive applications of waterfall. Curve elevation at each pixel corresponds to its grey value; shades of blue show updated pixel elevations after each image reconstruction; image boundaries and watershed lines are in red.}
\label{fig:waterfall}
\end{figure}

\section{GPU Waterfall Transform}
\label{sec:waterfall}

While the watershed transform is popular in many applications, its main drawback of over-segmentation is inherent to this procedure (Fig.~\ref{fig:qual}). Hierarchical segmentation relies on post-processing of the watershed transform to address the over-segmentation issue by merging non-significant catchment basins. The waterfall transform~\cite{beucher1994,marcotegui2005,golodetz2014two} constructs the adjacency graph of the watershed catchment basins and applies the watershed transform on the minimal spanning tree of this graph to achieve a new coarser segmentation of the input image. Recursive waterfall application results in a hierarchical segmentation of the image with coarser layers of fewer segmented regions produced at each iteration.

We propose a novel parallel waterfall reminiscent of the original~\cite{beucher1994} based on image reconstruction. After the primary watershed is applied on the input image, the image is modified by raising each of the catchment basins to a plateau the height of which equals the lowest pass point on the watershed line surrounding that catchment basin. In other words, the image modification comprises the two steps (kernels): V: identify new minima; VI: update image with new minima. Alg.~\ref{alg:newmin} provides the details of steps V--VI of the image reconstruction for the waterfall.

If $P$ is still the set of pixel positions in the input image, $I$ is the image value function, and $N(p)$ is the neighbourhood of the position $p \in P$, then for $p$ and $q \in N(p)$ on the watershed line, i.e.\ $L(p) \ne L(q)$, we define the watershed height as $height(p,q) = \max(I(p), I(q))$.

Since our watershed does not explicitly assign any pixel as belonging to watershed lines and all pixels are put into catchment basins, we can speculate that the watershed lines pass between pixels at the level of the higher pixels. When the pair $p$ and $q$ belong to different catchment basins, then $height(p, q)$ is the local height of the watershed line separating the catchment basins. For a given watershed catchment basin, its lowest pass point is the part of the surrounding watershed line with smallest height. This means that the height of the lowest pass point of a catchment basin equals the minimum among $height(p, q)$, where $p$ belongs to this catchment basin and $q$ does not. In the terrain map abstraction, if we flooded this catchment basin, then it would leak into a neighbouring catchment basin only after reaching the level, or the height, of the lowest pass point. Step V identifies this minimal level for each of the catchment basins by considering the whole watershed line. Before step V the array with the new minimum values ($newmin$) is initialised to a large constant $M$ such that $M$ is at least the upper bound of the greyscale range of the input image.

\begin{algorithm}[!htb]
\footnotesize
\caption{Image reconstruction for parallel waterfall}
\label{alg:newmin}
\begin{algorithmic}[1]
\STATE $memset(newmin, M, |P|)$ \COMMENT{Initialise new min to a large constant}
\FORALL[Step V: identify new min]{$p \in P$ in parallel} 
	\FORALL{$q \in N(p), L(q) \ne L(p)$}
		\STATE $height(p,q) \coloneq \max(I(p), I(q))$
		\STATE $newmin(L(p)) \coloneq \min(newmin(L(p)), height(p,q))$
	\ENDFOR
\ENDFOR
\FORALL[Step VI: update image with new min]{$p \in P$ in parallel}
	\IF{$I(p) < newmin(L(p))$}
		\STATE $I(p) \coloneq newmin(L(p))$
	\ENDIF
\ENDFOR
\end{algorithmic}
\end{algorithm}

Step VI updates the input image based on the new minimum values. Every pixel below the new minimum for its catchment basin is raised to the level of that new minimum, i.e.\ $I(p) = newmin(L(p))$ if, before step VI, $I(p)$ was smaller than $newmin(L(p))$.

Figure~\ref{fig:waterfall} illustrates the image reconstruction process for the waterfall. It depicts the watershed of an input 1D image followed by three recursive waterfall applications: the image reconstruction and the watershed. The image boundaries and watershed lines are red. The blue shades correspond to the new minima for each old watershed region. Each shade shows an image reconstruction to update pixel elevations.


If the watershed transform is applied on the reconstructed image, the result will be a set of new catchment basins, where each corresponds to several catchment basins in the primary watershed transform. The image reconstruction and the following watershed transform comprise the waterfall. If applied recursively, this transform gradually reduces the number of catchment basins, eventually grouping the whole image into a single region (Fig.~\ref{fig:waterfall}).

Our waterfall implementation relies on PRUF\textsubscript{bal} for the watershed steps I--IV. The image reconstruction steps V--VI are implemented with a single kernel each. Each pixel position in the image is still assigned to a separate execution thread; the parameters are set as per PRUF\textsubscript{bal}. The host code of the procedure that applies watershed followed by a fixed number of waterfall applications is presented in Alg.~\ref{alg:waterfall}.

\begin{algorithm}[!htb]
\footnotesize
\caption{The host code of the hierarchical segmentation}
\label{alg:waterfall}
\begin{algorithmic}[1]
\STATE $HostToDevice(I)$ \COMMENT{Copy image from host to device}
\STATE $DimGrid, DimBlock$ \COMMENT{Set grid and block dimensions}
\FOR{$n \in [0,\mathit{NL}-1]$}
	\STATE $InitialisationKernel$ \COMMENT{Step I}
        \STATE repeat $ResolutionKernel$ \COMMENT{Step II}
        \STATE repeat $ReductionKernel$ \COMMENT{Step III}
	\STATE $MergingKernel$	and $FullResolutionKernel$ \COMMENT{Step IV}
	\IF{$n < \mathit{NL} - 1$}
		\STATE $memset(newmin, M, |P|)$
		\STATE $IdentificationKernel$ \COMMENT{Step V}
		\STATE $UpdateKernel$ \COMMENT{Step VI}
	\ENDIF
	\STATE $DeviceToHost(L)$ \COMMENT{Copy labels from device to host}
\ENDFOR
\end{algorithmic}
\end{algorithm}

\begin{table*}[!htb]
\centering
\setlength{\tabcolsep}{4.5pt}
\begin{tabular}{c|c|c|*{6}{r}|r|r}
\multirow{2}{*}{\rotatebox{90}{GPU}} & \multirow{2}{*}{\rotatebox{90}{conn}} & source & \multicolumn{6}{c|}{C4L Image Dataset~\cite{c4l}} & \multicolumn{1}{c|}{MicroCT} & Berkeley~\cite{arbelaez2011} \\
\cline{3-11}
& & size & $256{\times}256$ & $512{\times}512$ & $1024{\times}1024$ & $1920{\times}1080$ & $2560{\times}1440$ & $2560{\times}1920$ & $4000{\times}4000$ & BSDS500 \\
\hline
\hline
\multirow{8}{*}{\rotatebox{90}{1080 (GPU-a)}} & \multirow{4}{*}{\rotatebox{90}{4}} & DW & 0.560 & 1.011 & 2.442 & 5.020 & 7.754 & 9.968 & 45.513 & 0.738 \\
& & PRW & 0.267 & 0.630 & 1.823 & 3.714 & 5.735 & 7.520 & 31.408 & 0.386 \\
& & APRUF & \textbf{0.224} & \textbf{0.572} & 1.910 & \textbf{3.528} & \textbf{5.593} & \textbf{7.379} & 65.215 & \textbf{0.337} \\
& & PRUF & 0.249 & 0.605 & \textbf{1.749} & 3.630 & 5.793 & 7.565 & \textbf{28.987} & 0.363 \\
\cline{2-11}
& \multirow{4}{*}{\rotatebox{90}{8}} & DW & 0.542 & 1.032 & 2.553 & 5.449 & 8.385 & 10.510 & 47.025 & 0.718 \\
& & PRW & 0.262 & 0.668 & 1.875 & 4.258 & 6.223 & 8.158 & 33.840 & 0.399 \\
& & APRUF & \textbf{0.221} & \textbf{0.613} & 1.890 & \textbf{4.096} & \textbf{6.169} & \textbf{8.041} & 67.851 & \textbf{0.360} \\
& & PRUF & 0.246 & 0.642 & \textbf{1.825} & 4.202 & 6.361 & 8.275 & \textbf{31.203} & 0.387 \\
\hline
\hline
\multirow{8}{*}{\rotatebox{90}{3050 Ti M (GPU-b)}} & \multirow{4}{*}{\rotatebox{90}{4}} & DW & 0.626 & 1.085 & 2.444 & 5.115 & 7.910 & 10.341 & 43.497 & 0.620 \\
& & PRW & 0.301 & 0.690 & 1.812 & 3.795 & 5.907 & 7.776 & 33.213 & 0.379 \\
& & APRUF & \textbf{0.264} & \textbf{0.615} & 1.816 & \textbf{3.657} & \textbf{5.701} & \textbf{7.530} & 56.532 & \textbf{0.328} \\
& & PRUF & 0.295 & 0.651 & \textbf{1.775} & 3.795 & 5.993 & 7.859 & \textbf{30.425} & 0.358 \\
\cline{2-11}
& \multirow{4}{*}{\rotatebox{90}{8}} & DW & 0.470 & 1.115 & 2.555 & 5.562 & 8.458 & 10.983 & 45.985 & 0.635 \\
& & PRW & 0.264 & 0.747 & 1.861 & 4.525 & 6.455 & 8.569 & 36.055 & 0.416 \\
& & APRUF & \textbf{0.212} & \textbf{0.685} & 1.847 & \textbf{4.395} & \textbf{6.341} & \textbf{8.270} & 72.390 & \textbf{0.372} \\
& & PRUF & 0.242 & 0.700 & \textbf{1.844} & 4.526 & 6.614 & 8.626 & \textbf{34.496} & 0.401 \\
\hline
\hline
\multirow{8}{*}{\rotatebox{90}{3060 Ti (GPU-c)}} & \multirow{4}{*}{\rotatebox{90}{4}} & DW & 1.133 & 1.592 & 2.802 & 4.925 & 7.451 & 9.502 & 34.430 & 1.310 \\
& & PRW & 0.531 & 0.791 & 1.647 & 3.039 & 4.660 & 6.046 & 21.659 & 0.635 \\
& & APRUF & \textbf{0.368} & \textbf{0.605} & \textbf{1.506} & \textbf{2.736} & \textbf{4.213} & \textbf{5.564} & 35.443 & \textbf{0.466} \\
& & PRUF & 0.457 & 0.717 & 1.517 & 2.885 & 4.448 & 5.794 & \textbf{19.775} & 0.549 \\
\cline{2-11}
& \multirow{4}{*}{\rotatebox{90}{8}} & DW & 1.162 & 1.610 & 3.013 & 5.203 & 7.785 & 9.462 & 35.501 & 1.317 \\
& & PRW & 0.537 & 0.843 & 1.804 & 3.495 & 5.036 & 6.202 & 23.351 & 0.668 \\
& & APRUF & \textbf{0.389} & \textbf{0.666} & \textbf{1.628} & \textbf{3.210} & \textbf{4.668} & \textbf{5.748} & 45.046 & \textbf{0.512} \\
& & PRUF & 0.470 & 0.769 & 1.674 & 3.359 & 4.894 & 5.991 & \textbf{21.820} & 0.593 \\
\end{tabular}
\caption{Different watershed algorithms on 2D images; execution times in ms. Boldface highlight is applied to the best performance.}
\label{tab:2Dres}
\end{table*}

After the image is copied into the device and the thread blocks are set up, the hierarchical segmentation is constructed layer by layer. The layer generation starts with a watershed run (lines 4--7). For all layers except the last one (if $0 \le n < \mathit{NL} - 1$) steps V--VI follow (lines 8--11) to prepare for the next watershed application by reconstructing the image with new minima levels. Finally, the new layer of the segmentation is copied from the device to the host memory (line~12).

The memory requirements of the hierarchical segmentation are the same as for PRUF\textsubscript{bal}: new variables or arrays are \emph{not} used. This is achieved by reusing the available memory: the same $I, L, S, S'$ arrays are used at every application of the watershed. This is allowed because: (a)~$L, S, S'$ are initialised from scratch during steps I--II and (b)~$L$ values are copied back into a new address in the host memory at each iteration. Furthermore, the $S'$ array is reused under the name $newmin$ in the image reconstruction steps V--VI.

The only new parameter that the hierarchical segmentation adds is the number of the waterfall applications. We denote the number of watershed applications (primary or as part of waterfall) $\mathit{NL}$, indicating the number of segmentation layers generated in the hierarchical segmentation. This parameter does not affect the execution of the individual waterfall applications. For a typical input image, a hierarchy leading to a single region corresponding to the whole image is usually achieved in under ten waterfall applications: $\mathit{NL}{=}5$ to $\mathit{NL}{=}7$ is usually enough to get a semantically meaningful hierarchy of segmentations. This estimate is based empirically on multiple previous experiments in our research group on the application of sequential watershed/waterfall to various data; results in Fig.~\ref{fig:qual} agree with this.

In terms of dimensionality, this hierarchical segmentation mimics the watershed transform used---PRUF\textsubscript{bal}, in this case. While a 3D implementation would work for 1D or 2D images, the memory consumption pattern of PRUF\textsubscript{bal} is optimised in case of separate 2D and 3D implementations.

\section{Evaluation and Results}
\label{sec:results}

\textbf{Datasets:} Two public datasets were used: C4L~\cite{c4l} (greyscale images grouped by resolution), and the Berkeley Segmentation Data Set and Benchmarks 500 (BSDS500)~\cite{arbelaez2011}, converted to greyscale.
We also used three 3D images (128 mega-voxels each: $320{\times}320{\times}1250$, $500{\times}500{\times}512$, $4000{\times}4000{\times}8$), and a 3D image of 800 megavoxels: $4000{\times}4000{\times}50$, constructed from the same microCT~\cite{yeghiazaryan2018path}. The $4000{\times}4000{\times}8$ image is also interpreted
as a set of eight 2D images. For reproducibility, all images are raw (no smoothing).


\textbf{Experimental Setup:} The experiments were carried out on three different machines to include processors and GPUs spanning multiple generations and compute capabilities: (GPU-a) Intel Core i7-8700, NVIDIA GeForce GTX 1080, CUDA 11.6, (GPU-b) Intel Core i9-12900HK, NVIDIA GeForce RTX 3050 Ti Mobile, CUDA 11.8, (GPU-c) AMD Ryzen 7 3700X, NVIDIA GeForce RTX 3060 Ti, CUDA 11.7. Every reported watershed execution time is a \emph{minimum} of ten runs per image (for reproducibility and to mitigate delays produced by other running processes), and an \emph{average} across multiple images in the same dataset (where applicable). To confirm the equivalence of different watershed procedures for a given image, we verify that they consistently generate the same number of regions. Although there are more powerful GPUs available in advanced supercomputer configurations, our selection demonstrates that the proposed approach is already feasible with standard computing hardware.

The three algorithms use two types of parameters: the block size for CUDA and the reduction rate \RR. The latter is only used in PRUF (Alg.~\ref{alg:pruf}), and PRW. 
Only execution time is affected by these parameters; otherwise, the results of different setups are equivalent. We have (partly) reported in~\cite{yeghiazaryan2018path,yeghiazaryan2018parallel} that CUDA block sizes of $16{\times}16$ in 2D and $8{\times}8{\times}8$ in 3D, and \RR$=6$ lead to fastest execution. 

\begin{figure}[!htb]
	\centering
	\includegraphics[width=.9\linewidth]{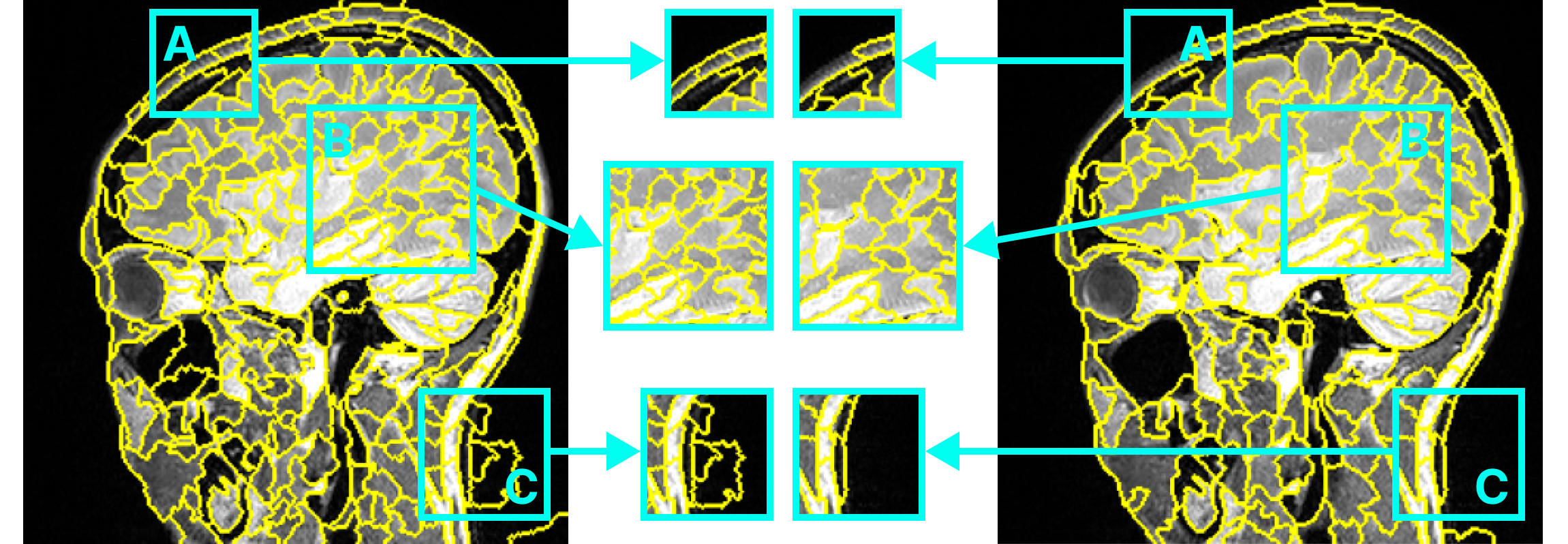}
	\caption{4- (left) vs 8-pixel (right) neighbourhood results on brain MRI, depicted as overlays with yellow waterfall layer 2 lines.}
\label{fig:brain_mri_w4vsw8}
\end{figure}

\subsection{Watershed Results in 2D and 3D}

Table~\ref{tab:2Dres} shows the comparison among PRUF, APRUF, PRW and DW~\cite{vitor2009,vitor2012analysis} algorithms. For all four algorithms 8-pixel and 4-pixel neighbourhoods were considered. 
%
%
For both C4L and BSDS500, APRUF (4-pixel on GPU-c) is the fastest: 0.37ms for the smallest image group ($256{\times}256$) and 5.56ms for the largest image group ($2560{\times}1920$).
APRUF's improvement over DW is between 41.44\% and 67.52\%.
PRUF (fastest) shows a 44.21\% improvement over APRUF (slowest) for the $4000{\times}4000$ images, indicating effectiveness of the reduction rate on larger images (detailed analysis in~\cite{gabrielyan2022}).
The comparison of the execution times among the 8-neighbour implementations of the algorithms shows a pattern similar to the case of 4-neighbour implementations discussed above. 
%



The average execution times of PRW, PRUF, APRUF, on 3D test images are reported in Table~\ref{tab:3Dres}. (We were unable to find 3D implementations of DW.) The $4000{\times}4000{\times}50$ image contains $8{\times}10^8$ voxels, which is too large for GTX 1080 and RTX 3050 Ti Mobile GPUs.

Across all large 3D images ($1.28{\times}10^8$ and $8{\times}10^8$ voxels) PRUF and PRW algorithms outperform APRUF, proving the effectiveness of the reduction rate on larger images. PRUF with 6-voxel connectivity (on GPU-c) outperforms PRW across all datasets, due to the parallel \union--\find\ implementation. 
In the best case ($500{\times}500{\times}512$) PRUF improves 30.77\% over APRUF.
Like in 2D, for small 3D images ($256{\times}256{\times}128$) APRUF outperforms the others.

Across the 128-megavoxel images, the noticeable difference in execution times
shows a correlation of the number of slices in the image and the execution time. Algorithms show longer execution times on images with more slices and smaller intra-slice resolutions due to the weaker locality.

\begin{table*}[!htb]
\centering
\setlength{\tabcolsep}{4.5pt}
\begin{tabular}{cc|c|c|r|*{3}{r}|r}
~&
\multirow{3}{*}{\rotatebox{90}{GPU}} & \multirow{3}{*}{\rotatebox{90}{conn}} & source & \multicolumn{1}{c|}{C4L} & \multicolumn{3}{c|}{MicroCT} & \multicolumn{1}{c}{MicroCT} \\
\cline{4-9}
& & & \# voxels & \multicolumn{1}{c|}{$8.39\times 10^6$} & \multicolumn{3}{c|}{$1.28\times 10^8$} & \multicolumn{1}{c}{$8{\times}10^8$} \\
\cline{4-9}
& & & volume & $256{\times}256{\times}128$ & $320{\times}320{\times}1250$ & $500{\times}500{\times}512$ & $4000{\times}4000{\times}8$ & $4000{\times}4000{\times}50$ \\
\hline
\hline
\multirow{6}{*}{\rotatebox{90}{GPU-a}} & 
\multirow{6}{*}{\rotatebox{90}{1080}} & 
\multirow{3}{*}{\rotatebox{90}{6}} & PRW & 17.061 & 359.172 & 303.864 & 303.116 & \multirow{3}{*}{GPU too small} \\
& & & APRUF & \textbf{15.617} & 582.493 & 435.270 & 605.604 & \\
& & & PRUF & 16.121 & \textbf{317.906} & \textbf{286.303} & \textbf{291.115} & \\
\cline{3-9}
& & \multirow{3}{*}{\rotatebox{90}{26}} & PRW & 35.299 & 853.416 & 695.573 & 598.303 & \multirow{3}{*}{GPU too small} \\
& & & APRUF & \textbf{33.613} & 935.466 & 754.057 & 913.332 & \\
& & & PRUF & 34.078 & \textbf{784.110} & \textbf{634.653} & \textbf{553.002} & \\
\hline
\hline
\multirow{6}{*}{\rotatebox{90}{GPU-b}} & 
\multirow{6}{*}{\rotatebox{90}{3050 Ti M }} & 
\multirow{3}{*}{\rotatebox{90}{6}} & PRW & 15.514 & 352.567 & 336.017 & 331.193 & \multirow{3}{*}{GPU too small} \\
& & & APRUF & \textbf{14.325} & 629.491 & 440.831 & 550.393 & \\
& & & PRUF & 15.486 & \textbf{323.475} & \textbf{319.391} & \textbf{322.818} & \\
\cline{3-9}
& & \multirow{3}{*}{\rotatebox{90}{26}} & PRW & 29.102 & 832.049 & \textbf{747.261} & \textbf{506.382} & \multirow{3}{*}{GPU too small} \\
& & & APRUF & \textbf{28.362} & 998.887 & 862.170 & 783.947 & \\
& & & PRUF & 29.048 & \textbf{807.878} & 748.010 & 511.280 & \\
\hline
\hline
\multirow{6}{*}{\rotatebox{90}{GPU-c}} & 
\multirow{6}{*}{\rotatebox{90}{3060 Ti}} & 
\multirow{3}{*}{\rotatebox{90}{6}} & PRW & 12.146 & 285.549 & 222.612 & 209.185 & 1438.270 \\
& & & APRUF & \textbf{10.944} & 471.418 & 300.419 & 330.670 & 2279.560 \\
& & & PRUF & 11.379 & \textbf{270.231} & \textbf{207.990} & \textbf{196.857} & \textbf{1357.750} \\
\cline{3-9}
& & \multirow{3}{*}{\rotatebox{90}{26}} & PRW & 21.283 & 565.747 & 460.407 & \textbf{315.526} & 2301.880 \\
& & & APRUF & \textbf{20.151} & 687.934 & 528.559 & 487.395 & 3506.960 \\
& & & PRUF & 20.549 & \textbf{559.511} & \textbf{453.554} & 315.927 & \textbf{2258.020} \\
\end{tabular}
\caption{Execution times (in ms) of the different watershed algorithms on 3D images. The best performance is in boldface.}
\label{tab:3Dres}
\end{table*}


\subsection{Effect of Connectivity Choice}

Table~\ref{tab:conn_catchemntbasins} analyses the number of catchment basins generated by 4- vs 8-pixel neighbourhood on different resolutions ($128^2$ to $4096^2$) of the same image~\cite{nasa_image}. 
In 2D, 4-pixel neighbourhoods produce 1.8($\pm$0.1) times more catchment basins than 8-pixel. The relative number of regions generated by each neighbourhood style is independent of the image resolution.

\begin{table}[!htb]
\centering
\setlength{\tabcolsep}{4.5pt}
\begin{tabular}{c|r*{5}r}

conn & 128 & 256 & 512 & 1024 & 2048 & 4096\\
\hline
\hline
4 & 1,748 & 6,523 & 24,217 & 98,107 & 273,728 & 413,901\\
8 & 926 & 3,458 & 13,196 & 52,191 & 165,869 & 241,572\\
\end{tabular}
\caption{Numbers of catchment basins for 4- and 8-pixel neighbourhoods for image resolutions $128^2$ to $4096^2$, on~\cite{nasa_image}.}
\label{tab:conn_catchemntbasins}
\end{table}

\begin{figure}[!htb]
    \centering
    \includegraphics[width=.9\linewidth]{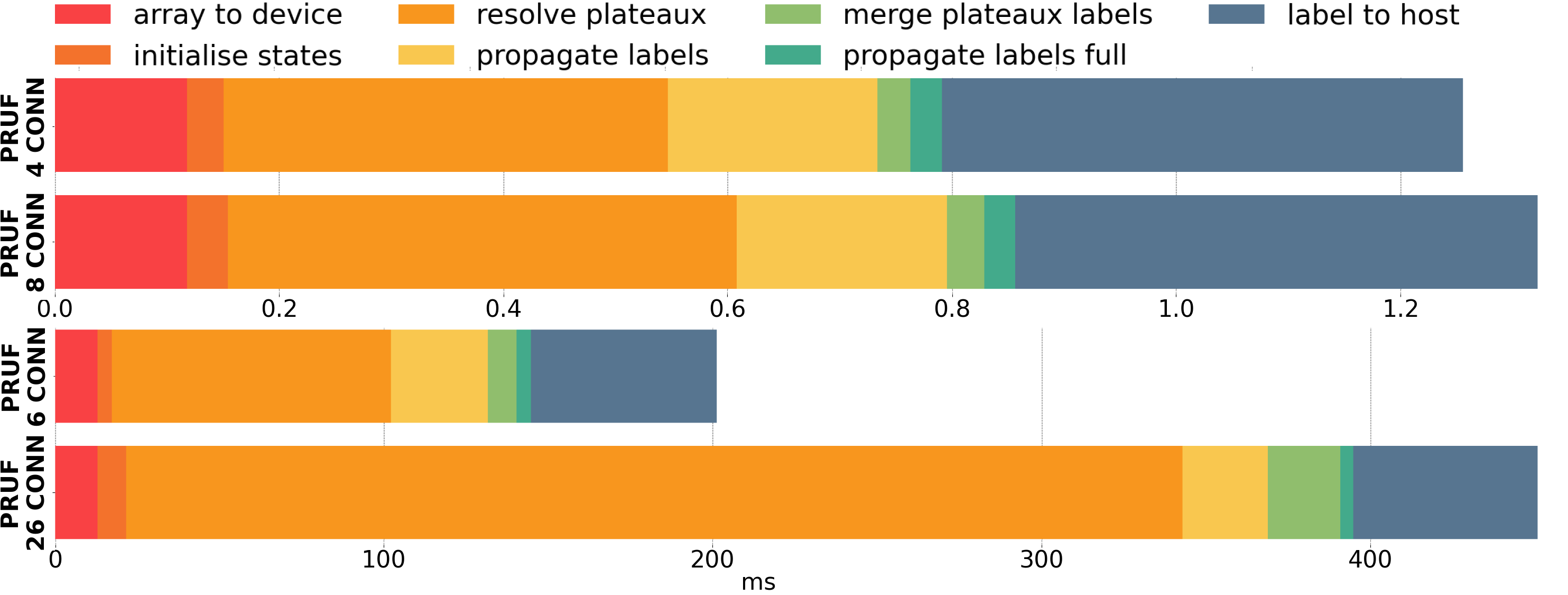}
    
    \caption{Composition of PRUF execution times for 2D and 3D Moore and von Neumann neighbourhoods: data copy,
 initialise states (I), resolve plateaux (II), propagate labels (III), merge plateaux and propagate labels (IV), data copy back.
    Measured on GPU-c using $1024{\times}1024$ (top) and $500{\times}500{\times}512$ (bottom) test images.}
    \label{fig:steps}
\end{figure}

Figure~\ref{fig:brain_mri_w4vsw8} shows the qualitative comparison between 4- and 8-pixel neighbourhoods on the waterfall layer~2. The 4-pixel neighbourhood tends to capture small details in the image (region A) but also sometimes over-segments regions (region C). Region B illustrates the difference in the number of catchment basins: 8-pixel version generates larger regions, hence it is slower (Fig.~\ref{fig:steps}).

\subsection{Waterfall Results}

\begin{figure}[!htb]
    \centering
    \includegraphics[width=.9\linewidth]{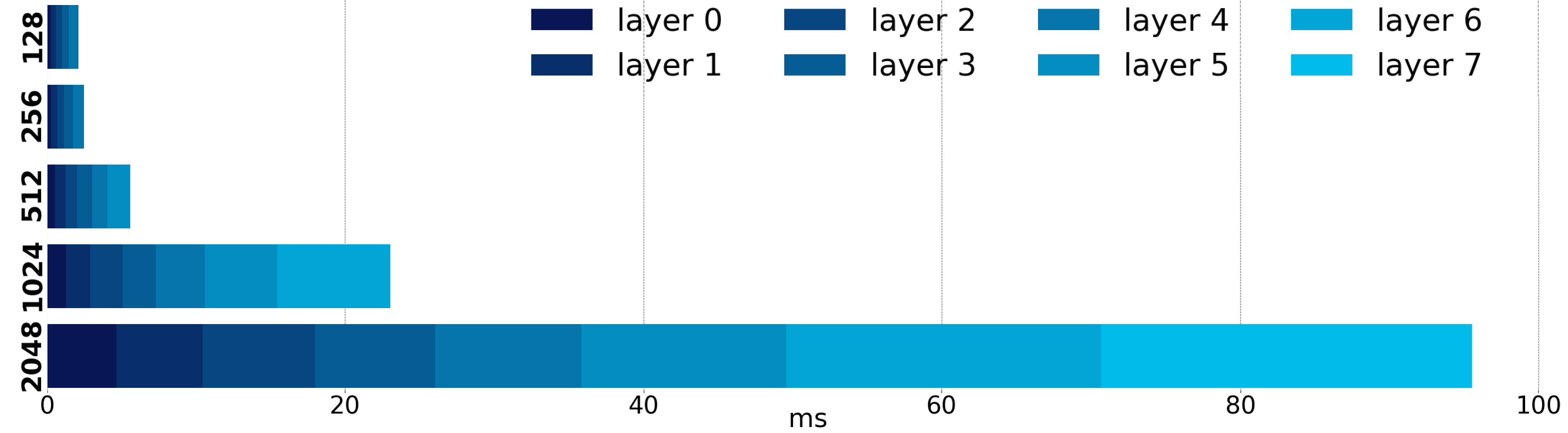}

    \caption{Execution time (in ms) of each layer of waterfall (PRUF) per image resolution ($128^2$ to $2048^2$), on the sample image~\cite{nasa_image} on GPU-c.}
    \label{fig:wf_nasa}
\end{figure}

Figure~\ref{fig:qual} illustrates the results of hierarchical segmentation with $\mathit{NL}{=}6$ on two 2D images: cameraman and knee MRI slice. The watershed (layer 0) and the waterfall transforms are applied on the gradient magnitude images of the smoothed input. The waterfall layers (1--5) demonstrate how smaller regions get gradually grouped together in higher layers of the segmentation. In each layer the remaining watershed lines trace the high-contrast edges in the image, thus moulding themselves around actual object boundaries. For instance, in segmentation layer 5 of the knee MRI slice, the femur is captured with a single waterfall region. For both images, segmentation layer 6 (not shown) contains only a single region corresponding to the whole image.

\begin{table}[!htb]
\centering
\setlength{\tabcolsep}{4.5pt}
\begin{tabular}{c|r*{5}r}

layer & 128 & 256 & 512 & 1024 & 2048 & 4096\\
\hline
\hline
0 & 1,748 & 6,523 & 24,217 & 98,107 & 273,728 & 413,901\\
1 & 277 & 1,089 & 3,703 & 13,808 & 41,233 & 56,164\\
2 & 44 & 194 & 578 & 2,158 & 6,230 & 8,652\\
3 & 10 & 34 & 94 & 320 & 952 & 1,334\\
4 & 2 & 6 & 16 & 50 & 138 & 202\\
5 & 1 & 1 & 4 & 10 & 21 & 29\\
6 & 1 & 1 & 1 & 3 & 5 & 5\\
7 & 1 & 1 & 1 & 1 & 2 & 2\\
8 & 1 & 1 & 1 & 1 & 1 & 1\\
\end{tabular}
\caption{Numbers of catchment basins in different waterfall layers (PRUF) in terms of image resolution ($128^2$ to $4096^2$), on~\cite{nasa_image}.}
\label{tab:wfcatchemntbasins}
\end{table}





Table~\ref{tab:wfcatchemntbasins} presents numbers of regions in different layers of the hierarchical segmentation on variants of the same image~\cite{nasa_image}.
%
Figure~\ref{fig:wf_nasa} compares layer generation runtimes using PRUF on~\cite{nasa_image} with different image resolution ($128^2$ to $2048^2$). Each measured layer generation time includes watershed with steps I--IV, image reconstruction with steps V--VI (except layer 7), and transfer of segmentation labels from device to host. The layer generation times increase with the layer number. 
The main runtime difference occurs in the duration of step II, which is due to larger non-minimal plateaux appearing after each image reconstruction step.

\begin{figure}[!htb]
    \centering
    \includegraphics[width=.9\linewidth]{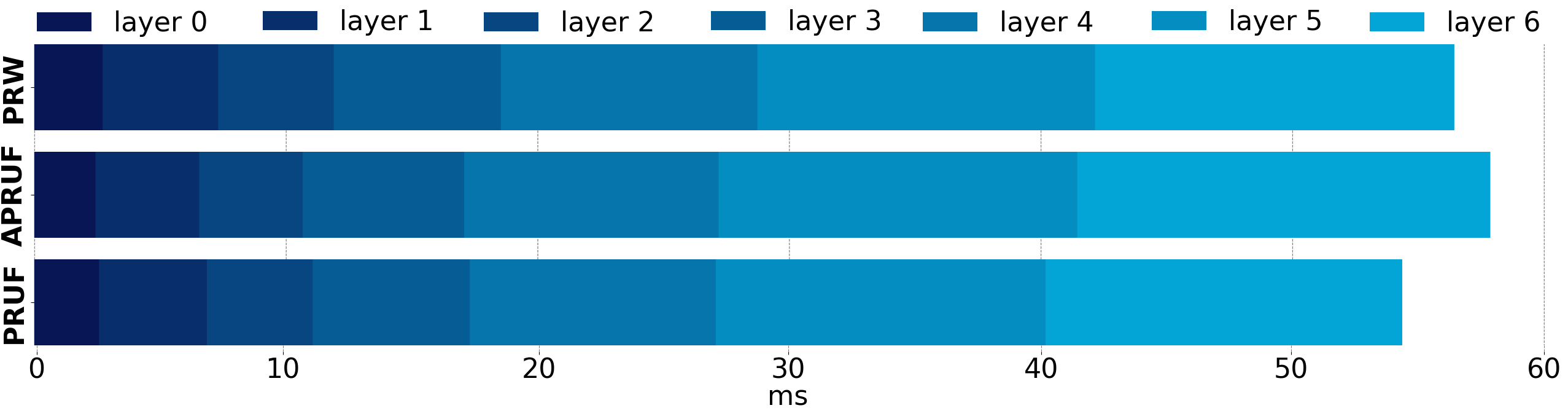}
    \caption{Comparison of mean execution times (in ms) of the waterfall versions: PRW, APRUF, PRUF on C4L $1024{\times}1024$ images on GPU-c.}
    \label{fig:wf_layer_1024}
\end{figure}

The comparison of average execution times (in ms) for the different waterfall algorithms for the C4L $1024{\times}1024$ test images is presented in Fig.~\ref{fig:wf_layer_1024}. As discussed, APRUF watershed (layer 0) is the fastest when it comes to smaller images. However, as PRUF is more efficient at handling label propagation for larger image regions, its execution becomes faster on further layers of waterfall. 

\begin{figure}[!htb]
	\centering
	\includegraphics[width=0.8\linewidth]{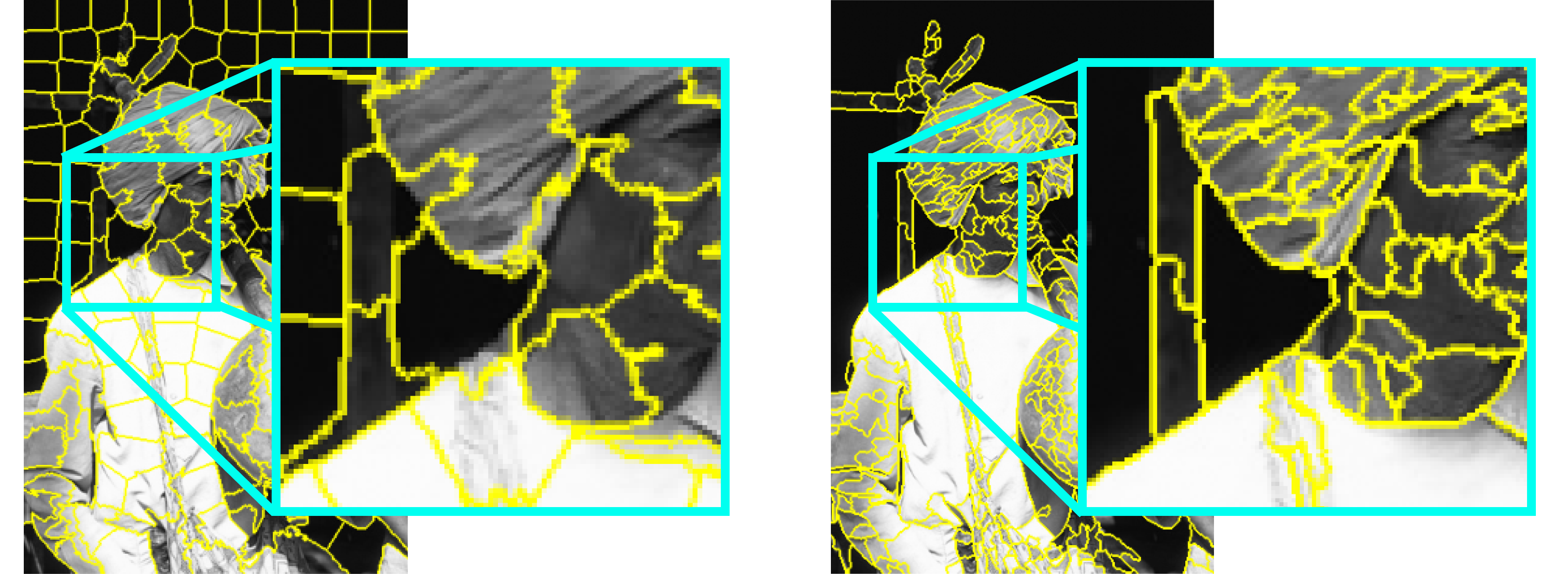}
	\caption{Edge preservation in SLIC (left) and waterfall layer $2$ (right).}
\label{fig:waterfall-slic}
\end{figure}

\section{Improving Semantic Segmentation Using APRUF}
\label{sec:super}

The sterile analysis in Sec.~\ref{sec:results} illustrates performance on isolated examples. Real-life Computer Vision applications of the waterfall algorithm are envisaged in circumstances where groups of pixels need to be decided upon in one go. The obvious choice of comparison is with superpixel-based algorithms~\cite{li2021superpixel}.
%
A superpixel is defined as a perceptually homogeneous region in the image. Usage of superpixel regions dramatically reduces the number of image primitives, thus improving the computational efficiency of the DL network. 

With the inherent regular shape and the grid-like distribution of segments, conventional superpixel algorithms ignore meaningful image edges by placing them inside a superpixel (Fig.~\ref{fig:waterfall-slic}). 
Since partitioning serves as a basis for later-stage processing, obtaining effective output is important, making watershed the preferred choice.

\begin{table*}[!h]
\centering
\setlength{\tabcolsep}{4.5pt}
\begin{tabular}{c|c|r*{3}{r}r|r}

Dataset & Metric & SLIC & FH & LSC & QS & SEEDS & Waterfall\\
\hline
\hline
\multirow{4}{*}{Indian Pines} & OA (\%) & 98.826$\pm$0.006 & 99.028$\pm$0.002 & 99.096$\pm$0.002 & 99.058$\pm$0.002 & 99.147$\pm$0.001 & \textbf{99.173$\pm$0.001}\\
& AA (\%) & 96.856$\pm$0.026 & 97.021$\pm$0.012 & 97.106$\pm$0.008 & 97.238$\pm$0.008 & \textbf{98.423$\pm$0.005} & 97.881$\pm$0.008\\
& $\kappa$ ($\times100$) & 98.662$\pm$0.007 & 98.891$\pm$0.002 & 98.969$\pm$0.002 & 98.927$\pm$0.002 & 99.027$\pm$0.001 & \textbf{99.057$\pm$0.001}\\
& time (s) & 7.866 & 7.904 & 7.995 & 8.213 & 7.776 & \textbf{7.744}\\
\hline
\multirow{4}{*}{University of Pavia} & OA (\%) & 99.582$\pm$0.001 & 99.508$\pm$0.001 & 99.432$\pm$0.002 & 99.379$\pm$0.002 & 99.546$\pm$0.001 & \textbf{99.596$\pm$0.001}\\
& AA (\%) & 99.371$\pm$0.002 & 99.241$\pm$0.002 & 99.060$\pm$0.003 & 99.026$\pm$0.003 & 99.343$\pm$0.002 & \textbf{99.387$\pm$0.003}\\
& $\kappa$ ($\times100$) & 99.447$\pm$0.001 & 99.347$\pm$0.001 & 99.247$\pm$0.003 & 99.177$\pm$0.003 & 99.398$\pm$0.002 & \textbf{99.464$\pm$0.002}\\
& time (s) & 104.038 & 83.181 & 89.200 & \textbf{81.097} & 87.892 & 83.477\\
\hline
\multirow{4}{*}{Salinas} & OA (\%) & 99.412$\pm$0.002 & 99.367$\pm$0.003 & 99.474$\pm$0.002 & 99.787$\pm$0.001 & 99.538$\pm$0.002 & \textbf{99.838$\pm$0.001}\\
& AA (\%) & 99.490$\pm$0.001 & 99.505$\pm$0.001 & 99.529$\pm$0.002 & 99.732$\pm$0.001 & 99.596$\pm$0.001 & \textbf{99.772$\pm$0.001}\\
& $\kappa$ ($\times100$) & 99.345$\pm$0.002 & 99.295$\pm$0.004 & 99.414$\pm$0.002 & 99.762$\pm$0.001 & 99.486$\pm$0.002 & \textbf{99.820$\pm$0.001} \\
& time (s) & 43.851 & 44.850 & 45.254 & 44.556 & 40.501 & \textbf{35.268}\\
\end{tabular}
\caption{Comparison of CEGCN with different superpixel segmentation methods vs waterfall method}
\label{tab:cegcn}
\end{table*}

\begin{table*}[!h]
	\centering
	\begin{tabular}{c|c|rrrr}
		Method & metric & ISIC skin & JSRT lung & JSRT heart & JSRT clavicle\\
	\hline\hline
	\multirow{2}{*}{SLIC} & DSC (\%) & \textbf{84.79$\pm$0.167} & 96.10$\pm$0.036 & 90.87$\pm$ 0.033 & \textbf{85.98$\pm$0.049}\\
	& time & \textbf{5h 43m 48s} & 10h 41m 46s & 10h 21m 47s & 6h \ 8m \ 5s\\
	\hline
	\multirow{2}{*}{Waterfall}& DSC (\%) & 83.07$\pm$0.166 & \textbf{96.41$\pm$0.033} & \textbf{91.61$\pm$0.044} & 85.86$\pm$0.049 \\
	& time & 10h 49m \ 8s & \textbf{3h 14m 13s} &\textbf{ 3h 11m 55s} & \textbf{1h 34m 15s}
	\end{tabular}
\caption{SLIC superpixel semantic segmentation~\cite{li2021superpixel} vs waterfall-based semantic segmentation on GPU-c}
\label{tab:noisymedseg}
\end{table*}


We compare the waterfall against superpixel using two DL tasks: hyperspectral image (HSI) classification and noisy label semantic segmentation.
CEGCN~\cite{liu2020cnn} is a heterogenous deep network that combines convolutional neural network (CNN) with graph convolutional network (GCN) for HSI classification. The CNN module is responsible for feature learning on small-scale regular regions, whereas the GCN, given the superpixel graph of the image, captures information from a larger-scale irregular structure.

In the setting of CEGCN, we compare waterfall against superpixel algorithms: SLIC~\cite{achanta2012slic}, quick shift~(QS)~\cite{vedaldi2008quick}, Felzenszwalb and Huttenlocher’s~(FH) method~\cite{felzenszwalb2004efficient}, linear spectral clustering~(LSC)~\cite{li2015superpixel}, and SEEDS~\cite{bergh2012seeds}. The comparison is conducted on three different HS images: Indian Pines, University of Pavia, and Salinas~\cite{liu2020cnn}. Each experiment is repeated ten times, reporting overall accuracy~(OA), average accuracy~(AA), and kappa coefficient~($\kappa$) and the training time. All the superpixel segmentation methods use the hyperparameters provided in~\cite{liu2020cnn}. The waterfall algorithm (4-connectivity with $\mathit{NL}{=}4$ for Indian Pines, University of Pavia datasets and 8-connectivity with $\mathit{NL}{=}3$ for Salinas dataset) is applied on the gradient magnitude of the compressed images after applying Gaussian blurring. The compressed image of the HSI is constructed by averaging values of all spectral bands. As proposed in~\cite{liu2020cnn}, superpixel algorithms are executed on the HSI prepossessed using linear discriminant analysis~(LDA). We found that skipping the preprocessing steps improved the results of the waterfall.
%
Table~\ref{tab:cegcn} reports that the waterfall outperforms the other algorithms.
The waterfall regions tend to be larger, thus they help speed up the training without loss of accuracy. 

A further set of experiments is conducted on the task of medical image segmentation from noisy labels~\cite{li2021superpixel}. The authors propose a robust iterative learning strategy that combines noise-aware training and noisy label refinement, all guided by superpixels. The design of the network mitigates the impact of noisy training labels by exploiting structural information. The model is built upon the assumption that pixels belonging to the same superpixel share similar ground truth labels. We believe that better object boundary preservation makes waterfall superior compared to superpixels. 

We compare waterfall to SLIC on the ISIC (skin) and JSRT (lung, heart, clavicle) datasets (cropped as per \cite{li2021superpixel}), with the mild label noise setting (noise ratio $\alpha{=}0.3$, noise level $\beta{=}0.5$). 
We evaluate the predicted masks with Dice similarity coefficient (DSC)~\cite{yeghiazaryan2018family} averaged over the last 10 epochs (out of 200).
%
Hyperparameters of SLIC are as described in~\cite{li2021superpixel}. Waterfall is applied on the gradient magnitude of the images with: 4-connecitivity for ISIC, JSRT lung, 8-connectivity for JSRT clavicle and heart; $\mathit{NL}{=}4$ for JSRT and $\mathit{NL}{=}2$ for ISIC. Training times and DSC scores are reported in Table~\ref{tab:noisymedseg}.

Waterfall shows DSC results comparable to SLIC, outperforming it on JSRT lung and heart datasets. Waterfall under-performance on the ISIC skin and JSRT clavicle data is due to the nature of the images/labels. Images in the ISIC skin dataset do not have clearly defined edges, making it more suitable for grid-like segmentation. A similar issue occurs on the JSRT clavicle dataset, where the given label does not have a clear edge separation on the images of the dataset. Waterfall, due to a smaller-sized segmentation graph, shows a significant improvement in the training time of the network, outperforming the SLIC model 3.9 times on the JSRT clavicle dataset. Note that for the ISIC skin dataset, with some loss of accuracy ($75.41\pm0.263\%$), the training time can be decreased by around 5 hours by increasing $\mathit{NL}{=}4$.


\section{Conclusion}
\label{sec:concl}

Our experiments show that in some DL applications, where edge preservation and fast training are important, choosing waterfall becomes more robust than using superpixels. 
%
%
%
To mitigate over-segmentation, we propose additional parallel steps that render our algorithms applicable iteratively as waterfall transforms, producing hierarchical segmentations of the image with fewer and larger regions. 
We show the practical applications of our waterfall algorithm as a replacement for superpixel algorithms in DL pipelines, and show consistent training time alongside accuracy improvement.

\bibliographystyle{elsarticle-num}
\bibliography{JPDC24-existingwork}

\end{document}